\documentclass[10pt,twocolumn,letterpaper]{article}

\usepackage[pagenumbers]{cvpr} 

\usepackage{graphicx}
\usepackage{amsmath}
\usepackage{amssymb}
\usepackage{booktabs}
\usepackage{caption}
\usepackage{subcaption}
\usepackage{mathtools}
\usepackage{floatrow}

\usepackage[pagebackref,breaklinks,colorlinks]{hyperref}

\usepackage[capitalize]{cleveref}
\crefname{section}{Sec.}{Secs.}
\Crefname{section}{Section}{Sections}
\Crefname{table}{Table}{Tables}
\crefname{table}{Tab.}{Tabs.}

\def\cvprPaperID{*****} 
\def\confName{CVPR}
\def\confYear{2022}

\begin{document}

\title{Accuracy Prediction for NAS Acceleration using Feature Selection and Extrapolation}

\author{Tal Hakim \\
Smart Shooter, Kibbutz Yagur, Israel \\
{\tt\small tal.hakim@smart-shooter.com}
}
\maketitle

\begin{abstract}
Predicting the accuracy of candidate neural architectures is an important capability of NAS-based solutions. When a candidate architecture has properties that are similar to other known architectures, the prediction task is rather straightforward using off-the-shelf regression algorithms. However, when a candidate architecture lies outside of the known space of architectures, a regression model has to perform extrapolated predictions, which is not only a challenging task, but also technically impossible using the most popular regression algorithm families, which are based on decision trees. In this work, we are trying to address two problems. The first one is improving regression accuracy using feature selection, whereas the other one is the evaluation of regression algorithms on extrapolating accuracy prediction tasks. We extend the NAAP-440 dataset~\cite{hakim2022naap} with new tabular features and introduce NAAP-440e, which we use for evaluation. We observe a dramatic improvement from the old baseline, namely, the new baseline requires 3x shorter training processes of candidate architectures, while maintaining the same mean-absolute-error and achieving almost 2x fewer monotonicity violations, compared to the old baseline's best reported performance. The extended dataset and code used in the study have been made public in the NAAP-440 repository~\footnote{\url{https://github.com/talcs/NAAP-440}}. 
\end{abstract}

\begin{figure}
     \centering
     \begin{subfigure}[b]{0.9\textwidth}
         \centering
         \includegraphics[width=\textwidth]{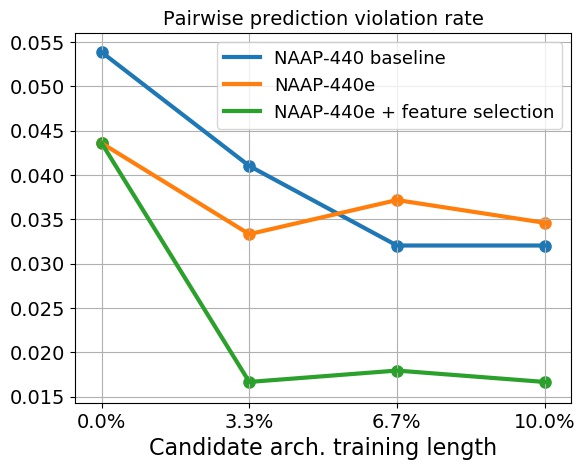}
     \end{subfigure}
     \begin{subfigure}[b]{0.9\textwidth}
         \centering
         \includegraphics[width=\textwidth]{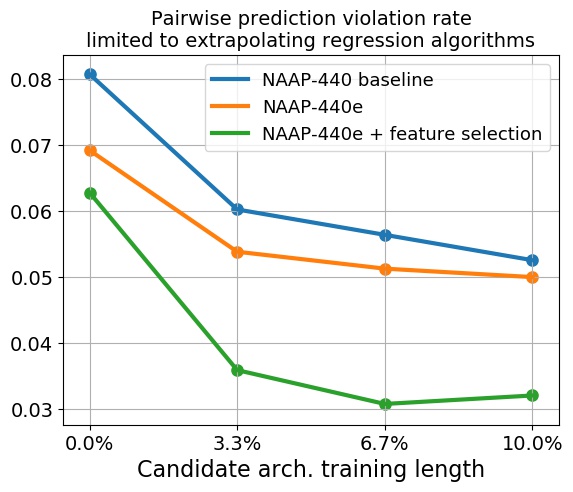}
     \end{subfigure}
    \caption{Performance improvements when adding new scheme features and applying feature selection, reporting the best performing regression algorithm of each baseline on each task.}
    \label{fig:predictionResults}
\end{figure}

\begin{figure}[]
\centering
  \includegraphics[width=\linewidth]{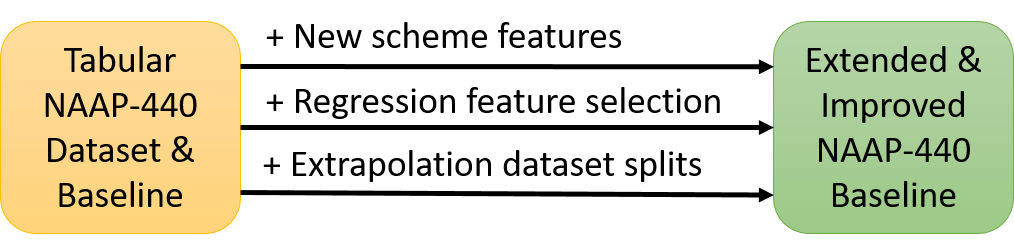}
  \caption[]{The NAAP-440 dataset is extended by adding new scheme features and new dataset splits for extrapolation. The regression baseline is dramatically improved when applying feature selection.}
  \label{fig:mainFigure}
\end{figure}

\section{Introduction}
\label{sec:intro}

Neural Architecture Search (NAS)~\cite{zoph2016neural} has been a dominating technique for proposing better performing neural architectures in recent years~\cite{zoph2018learning,tan2019efficientnet,tan2021efficientnetv2}. As a response to the high resource consumption of NAS and other similar approaches~\cite{zoph2018learning,liu2022convnet,radosavovic2020designing}, benchmark datasets have been proposed~\cite{ying2019bench,dong2020bench,baker2016designing,liu2018darts,dong2021nats,radosavovic2019network,hakim2022naap}, aiming to allow the evaluation of algorithms that can accelerate and improve the search process, by either using various search techniques or developing accuracy predictors, without actually training any candidate architecture~\cite{gracheva2021trainless,luo2020accuracy,wen2020neural,white2021powerful,xu2021renas,zela2020bench,siems2020bench,baker2017practical,istrate2019tapas,zhou2020performance,hakim2022naap,lukasik2020neural,lukasik2021smooth,friede2019variational,zela2022surrogate}.

Given a dense search space, predicting the accuracy of a candidate architecture that lies within the space is a rather simple task, using off-the-shelf regression algorithms that can be trained on the search space, such as Gradient Boosting, Random Forest and K-NN. However, when trying to solve more difficult problems, such as extrapolating from the training set search space, or mastering larger sets of features, there is room for improving the baseline results achieved by the off-the-shelf regression algorithms.

In this work, we aim to address the extrapolation and feature selection problems, as visualized in Figure~\ref{fig:mainFigure}. First, we introduce NAAP-440e, which is an extended version of the NAAP-440 tabular dataset~\cite{hakim2022naap}, which has new scheme features. Then, apart from the original train-test split of the dataset, we create three new splits, each imposing a different extrapolation problem. Finally, we apply a feature selection algorithm, to find an optimal feature set for each off-the-shelf regression algorithm on each dataset split. As visualized in Figure~\ref{fig:predictionResults}, we observe a tremendous improvement from the baseline by applying the described steps on the original dataset train-test split.

Our contributions in this work are as follows: (1) we introduce NAAP-440e, an extended version of the NAAP-440 dataset, which alone improves regression performances in many cases, (2) we advance the NAAP-440 baseline, by showing that feature selection can improve the accuracy prediction performances tremendously, especially when there are many features, (3) we map and report the most important features for the accuracy prediction problem, from the feature selection search graphs, (4) we introduce new baselines for the crucial problem of extrapolated predictions, which is essential when trying to compare candidate architectures that lie outside of the training set space.

\section{Advancement from NAAP-440 Baseline}
\label{sec:advancements}

\subsection{NAAP-440e: Additional Scheme Features}
\label{sec:scheme}
The NAAP-440 dataset originally contains 6 scheme features, which are the network’s depth, number of stages, first and last layer widths, number of total parameters and number of total MACs. We recognize two additional features that can be extracted from the scheme directly, which may be essential for the regression task.

\paragraph{Skip Connections.}
According to the NAAP-440 scheme generation, the third and fourth layers can have skip connections, as long as the requirement that a layer's input and output shapes match is satisfied. Even though each of the layers can independently have a skip connection, there is no architecture where both layers actually have a skip connection. This is caused by the fact that in order to allow multiple consecutive layers with skip connections, the input and output shapes of all the layers should match. In the NAAP-440 scheme generation, the third layer's input width can be either 16 or 24, whereas the fourth layer's output width can be either 32 or 40, such that the three shapes will never match. Therefore, the number of skip connections, which is one of the new scheme features we introduce, can only be assigned with one of the values $\{0,1\}$.

\paragraph{Lost Receptive Field.}
According to the NAAP-440 scheme generation, the kernel sizes of the third and fourth layers can be either 1x1 or 3x3, while their strides can be either 1 or 2. The case where the kernel size is 1x1 and the stride is 2 is not prevented. As a result, significant rates of the generated schemes will discard and lose some of the receptive field before reaching the classification stage. When the combination of 1x1 kernel with stride=2 is true for only one of the two layers, then 50\% of the receptive field is lost. When it is true for both, another 50\% of the receptive field is lost, such that the classification is only based on features from 25\% of the receptive field. Therefore, the second scheme feature we introduce is the number of layers where parts of the receptive field are lost, which can be assigned with one of the values $\{0,1,2\}$.

\subsection{Feature Selection}
\label{sec:featureSelection}
We implement feature selection to discover subsets of features that improve the accuracy of the regression algorithms. As described in NAAP-440, regression algorithms are evaluated using both MAE and a pairwise monotonicity score. Let us define the violation rate as

\begin{equation}
\label{eq:violationRate}
1 - MonotonicityScore = \frac{\#violations}{\binom{N}{2}}
\end{equation}

Given a regression algorithm, we would like to design a cost function to be minimized while we search for an optimal feature subset, which will take both MAE $m$ and violation rate $v$ into account. We check 3 possible cost functions, which are:

\begin{equation}
\begin{split}
m \cdot v, \\
m \cdot \log(v), \\
m \cdot \sqrt{v}.
\end{split}
\end{equation}

\noindent From the three options, we choose the cost function 
\begin{equation}
\label{eq:costFunction}
m \cdot \sqrt{v}, 
\end{equation}

\noindent as we find it best balancing between the two terms. The intuitions to not picking the other functions are: (1) we can see in Eq.~\ref{eq:violationRate} that minimizing the function $m \cdot v$ is equivalent to minimizing the function $m \cdot \#violations$, such that every single monotonicity violation adds another cost of $m$ to the cost function, which we find giving too much weight to the violation rate. (2) on the contrary, we can see from Eq.~\ref{eq:violationRate} that we can rewrite the cost function $m \cdot \log(v)$ as $m \cdot [\log(\#violations) - \log\binom{N}{2}]$, which is rather similar to minimizing $m \cdot \log(\#violations)$, where a solution with a slightly higher MAE will only be preferred in the case of a dramatic violation rate decrease, which we find as giving too low weight to the violation rate. We therefore prefer to use the cost function from Eq.~\ref{eq:costFunction}, which uses a square-root, whose violation rate's weight lies between those of the two other options. Finally, to boost the weight of the violation rate a little further, we refine the cost function to be

\begin{equation}
\label{eq:costFunctionFinal}
\mathrm{round}(m,3) \cdot \sqrt{v}.
\end{equation}

\paragraph{Feature Subset Search Space.}
\label{sec:featureSubsetSearchSpace}
Following the NAAP-440 baseline, we address the regression problem under 4 different conditions: (1) 100\% acceleration, where only scheme features are used for the regression, (2) 96.7\% acceleration, where features from the model's first 3 training epochs are added to the scheme features, (3) 93.3\% acceleration, where features from the first 6 training epochs are used, and (4) 90\% acceleration, where features from the first 9 training epochs are used. As the number of scheme features has now been increased from 6 to 8 and since each training epoch has three features, the number of features for each of the 4 conditions is 8, 17, 26, and 35, respectively. Given a set of of features $F$, the number of all possible feature subsets is $2^{|F|}$. For example, when using a set of 35 features, there are more than $3.2 \times 10^{10}$ possible subsets. Iterating over and evaluating all those subsets, aiming to find the optimal subset, is therefore infeasible for most of the regression algorithms, whose training and evaluation last more than a few milliseconds. 

\paragraph{Search Strategy.}
Aiming to find a satisfactory result in a practical computation time, we implement a variation of the hill climbing optimization algorithm with BFS. We initialize a priority queue with a single, initial state, which is the set of all features. At each step, we dequeue from the priority queue the list of states that share the best score. Each of the dequeued states, which represents a unique subset of features, is evaluated on a given regression algorithm. Then, its immediate neighboring states are computed and are added to the priority queue with the evaluation score of the current state. This technique allows us to stop the search at a rather early stage and still achieve a satisfactory result, which is usually dramatically better than the initial state of all features.

\paragraph{Stopping Criteria.}
While running our hill climbing optimization, we always use our priority queue to query the most promising next states to analyze, even if we are currently at a local minima, since we like to avoid stopping at an early suboptimal local minima. Therefore, we define a stopping criteria that only relates to the number of descent steps taken so far. For effective results, the stopping criteria should be aligned with the size of the feature set. For example, if the optimal subset contains 4 out of 8 features, then it will take at least 4 descent steps to discover it, since our initial state contains all 8 features, while an immediate neighbor discovered at each step has exactly one mismatch with the current state. More generally, given that the rate $r$ of useless features is constant, it will require at least $r|F|$ descent steps to find the optimal feature subset. Therefore, we find it essential to define a stopping criteria that has a linear relationship with the number of optimized parameters, which is the number of features. So, we define the number of descent steps to be $p|F|$, where $p$ is a parameter we set to $1$ in our experiments.

\paragraph{Stochastic Pruning.}
\label{sec:stochasticPruning}
When performing regression using the K-NN algorithm, especially when having many features, there is a non-negligible chance that multiple feature subsets $S$ will achieve the same score, causing the addition of $|S| \cdot |F|$ candidate subsets to the priority queue with the same score. That is because removing or adding features does not necessarily affect the identity of the K nearest neighbors of the test samples, meaning that the predictions will remain unchanged and thus multiple subsets will achieve the same score. This means that with no pruning, some of the descent steps might include the analysis of a very high amount of feature subsets. As more candidate subsets are evaluated at a single descent step, the chance of achieving identical scores by different subsets grows, whereas the number of discovered neighbors grows too, such that the phenomenon is amplified when we search deeper. To address it, we limit the number of evaluated subsets at each descent step to $b \cdot |F|$, where $b$ is the max branching factor, which we set as 3 in our experiments. In practice, when the current descent step has more than $b \cdot |F|$ feature subsets to evaluate, we only check $b \cdot |F|$ randomly sampled subsets, while returning the others to the queue. 

\begin{figure}[]
\centering
  \includegraphics[width=\linewidth]{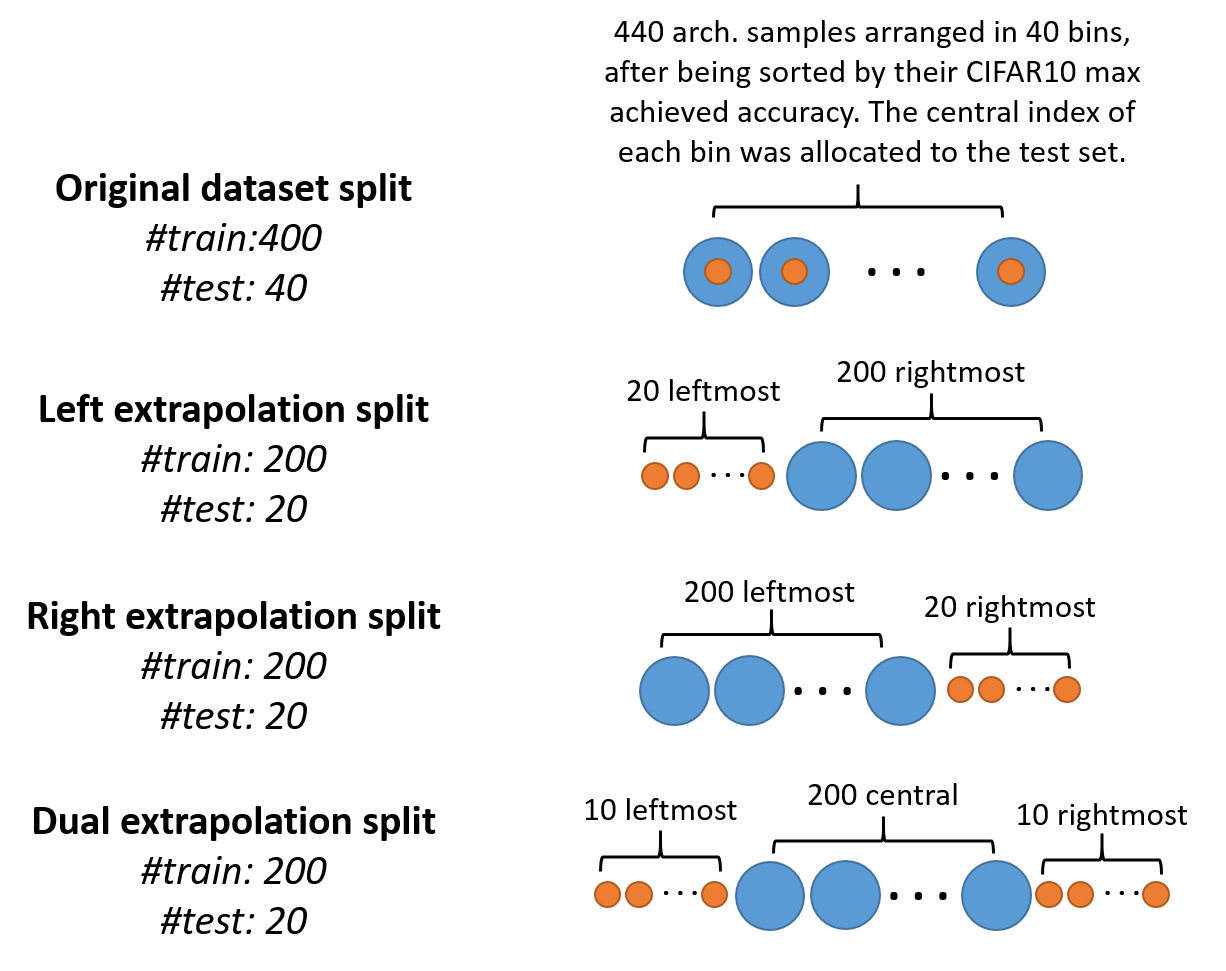}
  \caption[]{We apply 3 new splits on the already-split NAAP-440e dataset, which allow the evaluation of regression algorithm on extrapolation tasks.}
  \label{fig:datasetSplits}
\end{figure}

\subsection{Extrapolation}
\label{sec:extrapolation}

\paragraph{Linear Regression with Nonlinear Activation.}
As the baseline's best performing regression algorithms cannot extrapolate, we aim at extending the list of regression algorithms that can extrapolate. The linear regression least-squares optimization finds an optimal weight vector $\vec{w}$ such that $\sum_i(\vec{x_i} \vec{w} - y_i)^2$ is minimized, given feature vectors $x_i$ and their corresponding ground truth values $y_i$. That way, given a test sample $\vec{x}$, we can multiply it by the optimal weight vector $\vec{w}$ and get the approximation of its corresponding $y$ based on our training data. We would now like to apply a nonlinear activation function $f$ on the inner product $\vec{x}\vec{w}$ and solve for $\vec{w}$ such that $\sum_i(f(\vec{x_i} \vec{w}) - y_i)^2$ is minimized. Since we treat our regression algorithm as a black-box that can only handle linear cases, we approach this problem by applying the inverse activation function $f^{-1}$ on both sides. This gives us $\sum_i(f^{-1}(f(\vec{x_i} \vec{w})) - f^{-1}(y_i))^2$, which is simplified to $\sum_i(\vec{x_i} \vec{w} - f^{-1}(y_i))^2$. In other words, we apply the inverse nonlinear function $f^{-1}$ on the ground truth values and apply linear regression on it. This means that given a test sample $\vec{x}$ and it ground truth $y$, we will need to compute $f(\vec{x}\vec{w})$ in order to approximate $y$. The nonlinear activation functions $f$ that we check are:

\begin{equation}
\begin{split}
f = (x + 1)^\frac{1}{2}, \\
f = (x + 1)^\frac{1}{4}, \\
f = (x + 1)^2, \\
f = e^x \\
f = \log(x) \\
f = \frac{1}{1 + e^{-x}} \\
\end{split}
\end{equation}

\begin{figure}
    \RawFloats
     \centering
     \begin{subfigure}[b]{0.48\textwidth}
         \centering
         \includegraphics[width=\textwidth]{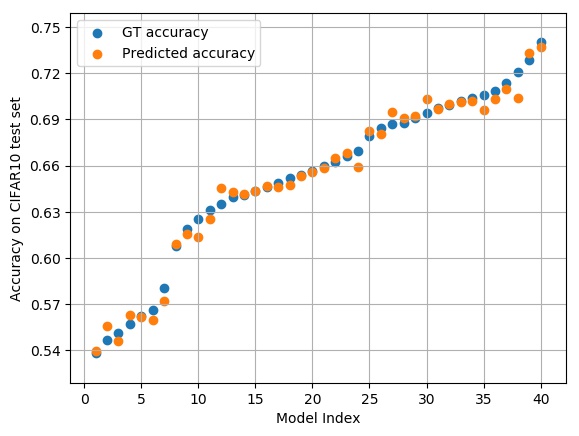}
     \end{subfigure}
     \hfill
     \begin{subfigure}[b]{0.48\textwidth}
         \centering
         \includegraphics[width=\textwidth]{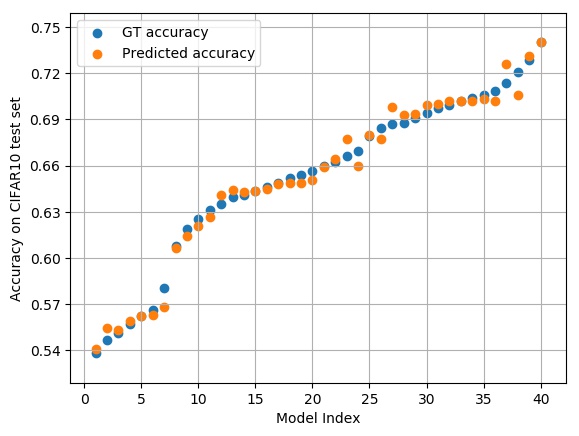}
     \end{subfigure}

     \begin{subfigure}[b]{0.48\textwidth}
         \centering
         \includegraphics[width=\textwidth]{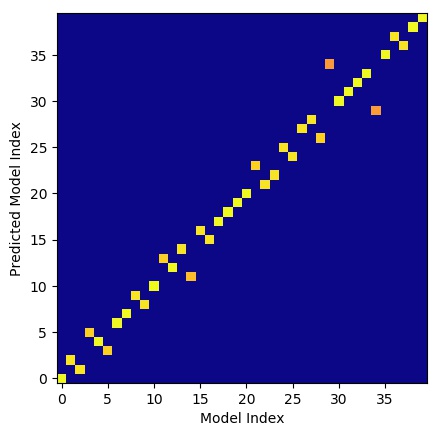}
         \caption{Old baseline best performance: Random Forest (N=200 estimators) with features from first 9 training epochs (25 monotonicity violations).}
     \end{subfigure}
     \hfill
     \begin{subfigure}[b]{0.48\textwidth}
         \centering
         \includegraphics[width=\textwidth]{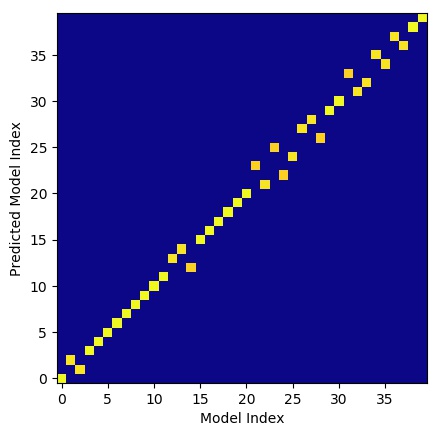}
         \caption{New baseline best performance: Gradient Boosting (N=200 estimators) with features from first 3 training epochs (13 monotonicity violations).}
     \end{subfigure}
     
        \caption{Old and new baselines best regression performances.}
        \label{fig:oldVsNewBaseline}
\end{figure}

\begin{figure*}
     \centering
     \begin{subfigure}[b]{0.24\textwidth}
         \centering
         \includegraphics[width=\textwidth]{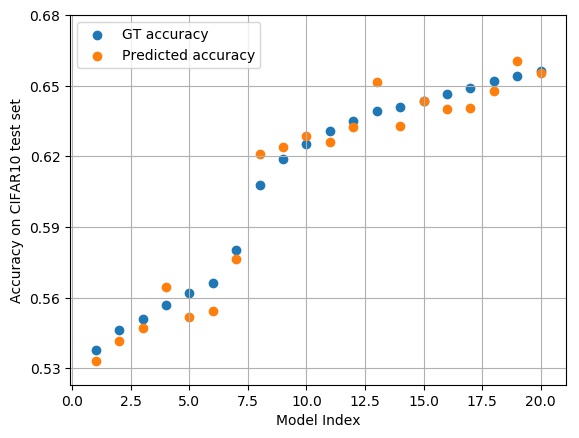}
     \end{subfigure}
     \hfill
     \begin{subfigure}[b]{0.24\textwidth}
         \centering
         \includegraphics[width=\textwidth]{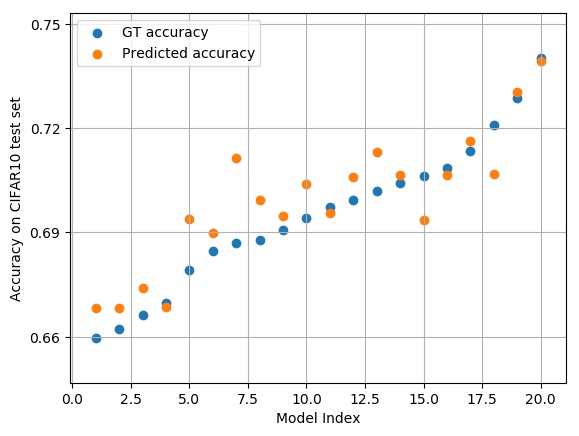}
     \end{subfigure}
     \hfill
     \begin{subfigure}[b]{0.24\textwidth}
         \centering
         \includegraphics[width=\textwidth]{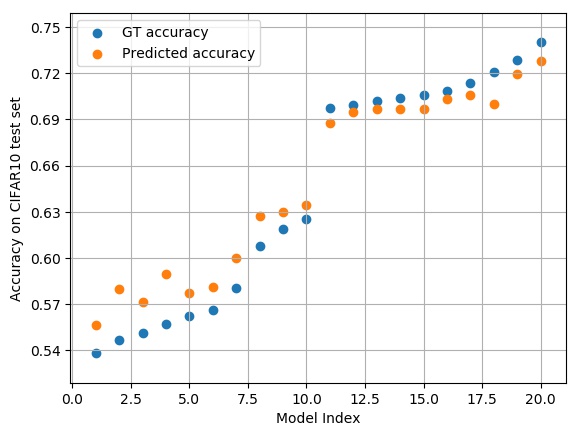}
     \end{subfigure}
     \hfill
     \begin{subfigure}[b]{0.24\textwidth}
         \centering
         \includegraphics[width=\textwidth]{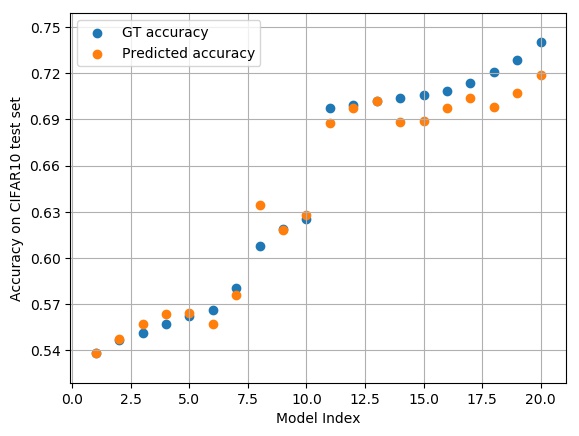}
     \end{subfigure}
     
     \begin{subfigure}[b]{0.24\textwidth}
         \centering
         \includegraphics[width=\textwidth]{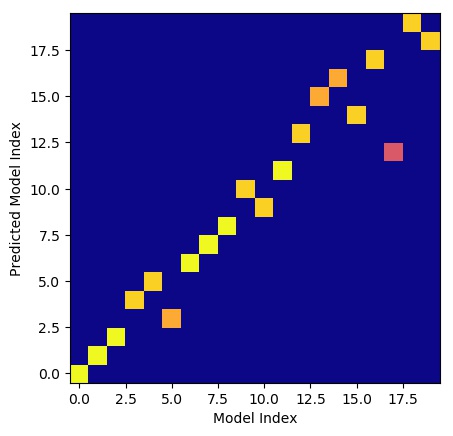}
         \caption{Left extrapolation performance with Linear Regression (D=0.25) and features from 9 training epochs. }
     \end{subfigure}
     \hfill
     \begin{subfigure}[b]{0.24\textwidth}
         \centering
         \includegraphics[width=\textwidth]{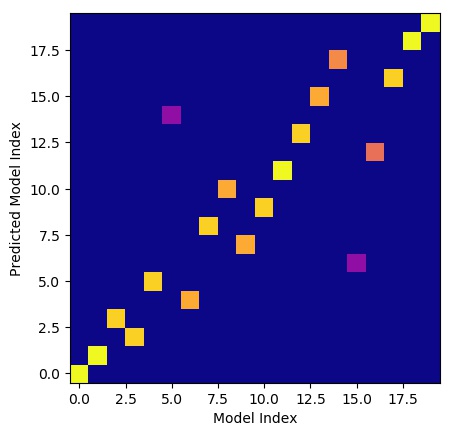}
         \caption{Right extrapolation performance with Linear Regression (D=2) and features from 9 training epochs.}
     \end{subfigure}
     \hfill
     \begin{subfigure}[b]{0.24\textwidth}
         \centering
         \includegraphics[width=\textwidth]{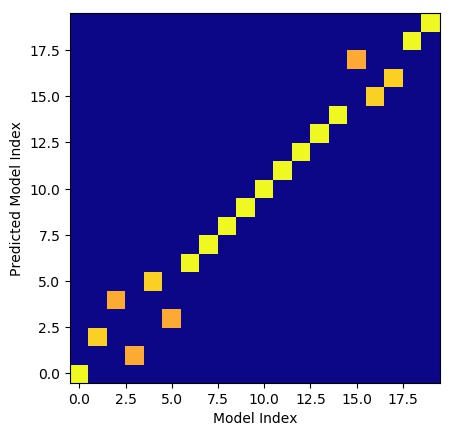}
         \caption{Dual extrapolation performance with Linear Regression (D=0.25) and features from 6 training epochs.}
     \end{subfigure}
     \hfill
     \begin{subfigure}[b]{0.24\textwidth}
         \centering
         \includegraphics[width=\textwidth]{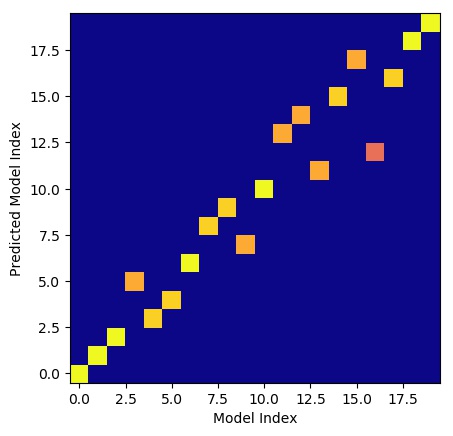}
         \caption{Dual extrapolation performance with Linear Regression (D=0.25) and features from 9 training epochs.}
     \end{subfigure}
        \caption{Visualized performances on the extrapolation splits.}
        \label{fig:extrapolations}
\end{figure*}

\paragraph{Dataset Splits.}
We would like to create new dataset splits that will impose extrapolation problems. As visualized in Figure~\ref{fig:datasetSplits}, the original split is uniform, since the 440 samples are first sorted by their GT accuracy and then arranged in 40 bins, each containing 11 samples. Then, for each bin, the 11 samples are split such that the sample at the center is allocated for the test set, while the other 10 samples are allocated for the training set. We keep using the same 40 bins when creating the three new splits. For example, the split for left extrapolation problem is achieved by neglecting the 200 leftmost training samples, which all belong to the 20 leftmost bins, while also neglecting the 20 rightmost test samples, which all belong to the 20 other bins. Likewise, as illustrated in Figure~\ref{fig:datasetSplits}, we create a split for the right extrapolating problem, by neglecting the training samples from the 20 rightmost bins and the test samples from the 20 leftmost bins. The third split is designated for extrapolating to both sides, as it neglects the training samples from the 10 leftmost and 10 rightmost bins, as well as the test samples from the 20 central bins. As a result, while the original split contains 400 training samples and 40 test samples, each of the new splits contains 200 training samples and 20 test samples. The choice of using the existing bins for the new splits rather than rearranging the data is based on the fact that the GT accuracies of the architectures are very dense, such that if our test set contains architectures from the same bin, it will be almost impossible to compare them, such that their monotonicity score will become meaningless.

\begin{figure*}[]
\centering
  \includegraphics[width=\linewidth]{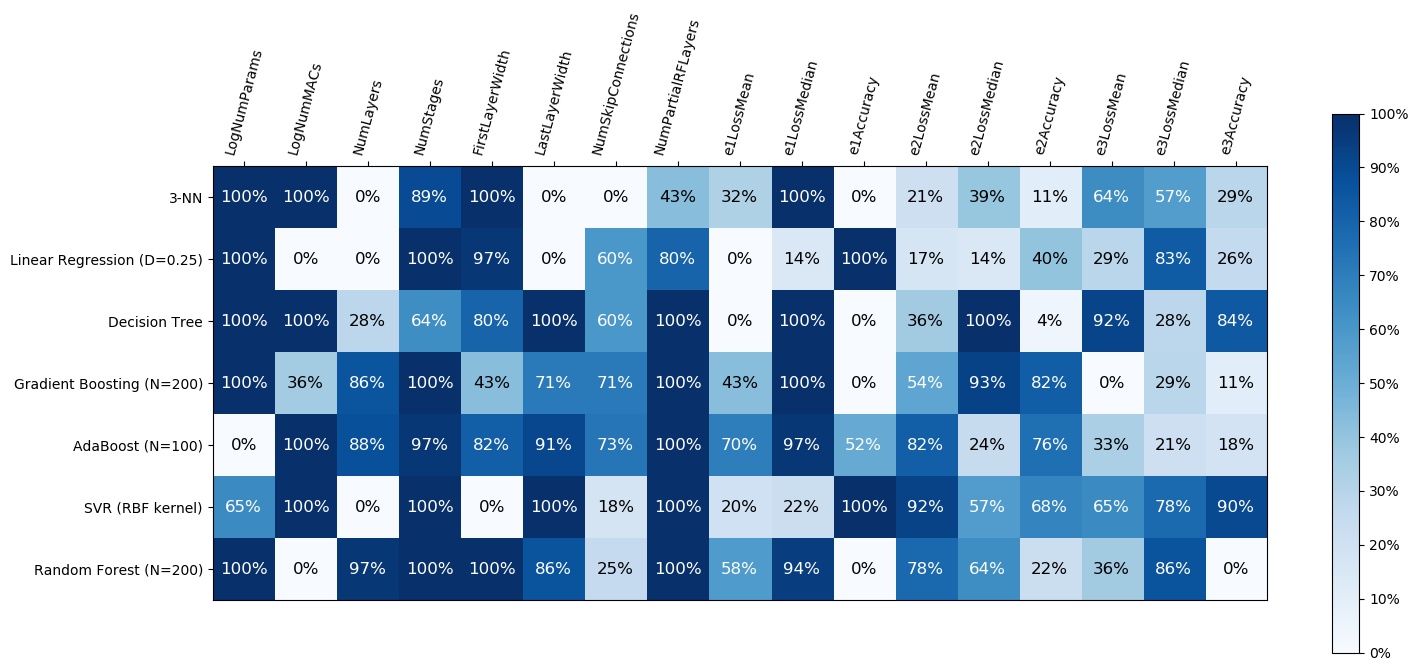}
  \caption[]{Feature selection statistics on 8\% best feature subsets per regression algorithm, when using scheme features and 3 first epoch features.}
  \label{fig:selectedFeaturesSchemeAndQuant3}
\end{figure*}

\begin{figure}[]
\centering
  \includegraphics[width=\linewidth]{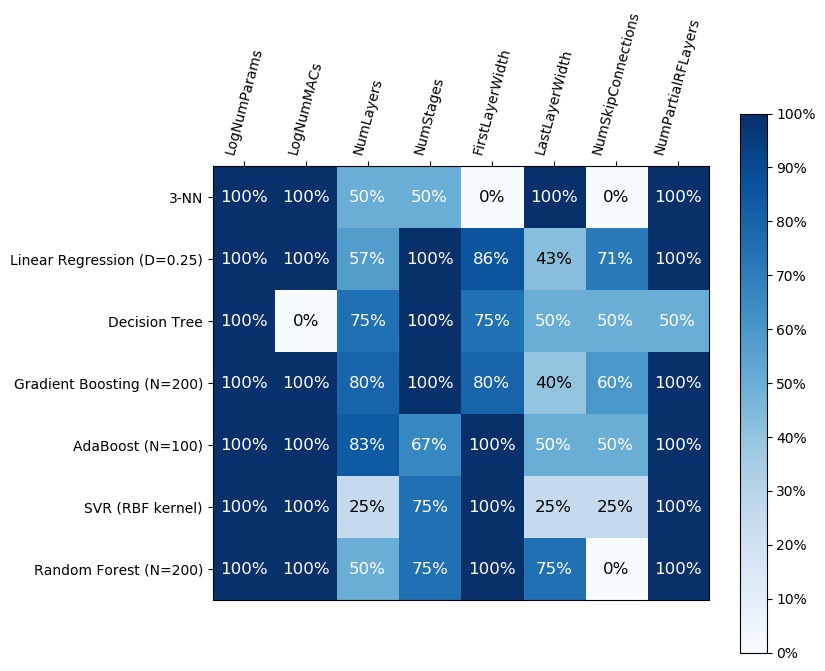}
  \caption[]{Feature selection statistics on 8\% best feature subsets per regression algorithm, when only using scheme features.}
  \label{fig:selectedFeaturesScheme}
\end{figure}

\begin{table*}[]
\centering
\resizebox{0.83\linewidth}{!}{%
\begin{tabular}{l||c|c|c|c}
\hline
\hline
 &  \multicolumn{4}{c}{\textbf{MAE / Monotonicity Score / \#Monotonicity Violations}} \\
 \hline
 & 100.0\% acceleration & 96.7\% acceleration & 93.3\% acceleration & 90.0\% acceleration\\ 
\textbf{Algorithm} & (0 epochs) & (3 epochs) & (6 epochs) & (9 epochs)\\ 
\hline
3-NN & 0.011 / 0.908 / 72  & 0.008 / 0.933 / 52  & 0.008 / 0.944 / 44  & 0.008 / 0.942 / 45 \\ 
 & 0.009 / 0.938 / 48  & 0.004 / 0.972 / 22  & 0.005 / 0.982 / 14  & 0.004 / 0.974 / 20 \\ 
\hline
Linear Regression (D=0.25) & 0.011 / 0.931 / 54  & 0.008 / 0.944 / 44  & 0.007 / 0.944 / 44  & 0.006 / 0.947 / 41 \\ 
 & 0.011 / 0.932 / 53  & 0.006 / 0.954 / 36  & 0.006 / 0.962 / 30  & 0.006 / 0.968 / 25 \\ 
\hline
Decision Tree & 0.009 / 0.921 / 62  & 0.007 / 0.927 / 57  & 0.007 / 0.924 / 59  & 0.007 / 0.929 / 55 \\ 
 & 0.007 / 0.936 / 50  & 0.005 / 0.964 / 28  & 0.004 / 0.972 / 22  & 0.004 / 0.979 / 16 \\ 
\hline
Gradient Boosting (N=200) & 0.005 / 0.956 / 34  & 0.006 / 0.960 / 31  & 0.006 / 0.962 / 30  & 0.005 / 0.949 / 40 \\ 
 & 0.005 / 0.956 / 34  & 0.004 / 0.983 / 13  & 0.004 / 0.977 / 18  & 0.004 / 0.983 / 13 \\ 
\hline
AdaBoost (N=100) & 0.009 / 0.918 / 64  & 0.007 / 0.950 / 39  & 0.006 / 0.954 / 36  & 0.006 / 0.956 / 34 \\ 
 & 0.008 / 0.923 / 60  & 0.006 / 0.964 / 28  & 0.006 / 0.959 / 32  & 0.004 / 0.963 / 29 \\ 
\hline
SVR (RBF kernel) & 0.010 / 0.926 / 58  & 0.007 / 0.938 / 48  & 0.006 / 0.962 / 30  & 0.005 / 0.958 / 33 \\ 
 & 0.008 / 0.936 / 50  & 0.006 / 0.967 / 26  & 0.004 / 0.977 / 18  & 0.004 / 0.978 / 17 \\ 
\hline
Random Forest (N=200) & 0.007 / 0.944 / 44  & 0.005 / 0.959 / 32  & 0.005 / 0.963 / 29  & 0.005 / 0.965 / 27 \\ 
 & 0.006 / 0.951 / 38  & 0.004 / 0.976 / 19  & 0.005 / 0.972 / 22  & 0.004 / 0.978 / 17 \\ 
\hline

\hline
\end{tabular}
}
\caption[]{The improvement caused by applying feature selection. At each cell, the upper row reports the performances with all features included, while the lower row reports the performance with the best feature subset found.}
\label{tbl:allFeaturesVsBestFeatures}
\end{table*}

\begin{table*}[]
\centering
\resizebox{0.83\linewidth}{!}{%
\begin{tabular}{l||c|c|c|c}
\hline
\hline
 &  \multicolumn{4}{c}{\textbf{MAE / Monotonicity Score / \#Monotonicity Violations}} \\
 \hline
 & 100.0\% acceleration & 96.7\% acceleration & 93.3\% acceleration & 90.0\% acceleration\\ 
\textbf{Algorithm} & (0 epochs) & (3 epochs) & (6 epochs) & (9 epochs)\\ 
\hline
1-NN & 0.007 / 0.928 / 56  & 0.004 / 0.967 / 26  & 0.003 / 0.982 / 14  & 0.005 / 0.967 / 26 \\ 
3-NN & 0.009 / 0.938 / 48  & 0.004 / 0.972 / 22  & 0.005 / 0.982 / 14  & 0.004 / 0.974 / 20 \\ 
5-NN & 0.008 / 0.938 / 48  & 0.004 / 0.965 / 27  & 0.005 / 0.978 / 17  & 0.005 / 0.974 / 20 \\ 
7-NN & 0.008 / 0.940 / 47  & 0.005 / 0.973 / 21  & 0.005 / 0.978 / 17  & 0.006 / 0.969 / 24 \\ 
9-NN & 0.008 / 0.935 / 51  & 0.005 / 0.971 / 23  & 0.005 / 0.978 / 17  & 0.006 / 0.968 / 25 \\ 
Linear Regression & 0.014 / 0.926 / 58  & 0.008 / 0.953 / 37  & 0.007 / 0.962 / 30  & 0.006 / 0.965 / 27 \\ 
Linear Regression (D=0.5) & 0.013 / 0.928 / 56  & 0.007 / 0.954 / 36  & 0.006 / 0.967 / 26  & 0.005 / 0.963 / 29 \\ 
Linear Regression (D=0.25) & 0.011 / 0.932 / 53  & 0.006 / 0.954 / 36  & 0.006 / 0.962 / 30  & 0.006 / 0.968 / 25 \\ 
Linear Regression (D=2) & 0.014 / 0.924 / 59  & 0.008 / 0.949 / 40  & 0.007 / 0.959 / 32  & 0.006 / 0.962 / 30 \\ 
Linear Regression (Exp) & 0.014 / 0.915 / 66  & 0.009 / 0.951 / 38  & 0.008 / 0.969 / 24  & 0.007 / 0.959 / 32 \\ 
Linear Regression (Log) & 0.013 / 0.928 / 56  & 0.007 / 0.951 / 38  & 0.006 / 0.967 / 26  & 0.006 / 0.967 / 26 \\ 
Linear Regression (Sigmoid) & 0.013 / 0.928 / 56  & 0.007 / 0.953 / 37  & 0.007 / 0.962 / 30  & 0.006 / 0.967 / 26 \\ 
Decision Tree & 0.007 / 0.936 / 50  & 0.005 / 0.964 / 28  & 0.004 / 0.972 / 22  & 0.004 / 0.979 / 16 \\ 
Gradient Boosting (N=25) & 0.007 / 0.937 / 49  & 0.006 / 0.967 / 26  & 0.005 / 0.969 / 24  & 0.005 / 0.969 / 24 \\ 
Gradient Boosting (N=50) & 0.006 / 0.949 / 40  & 0.005 / 0.974 / 20  & 0.004 / 0.972 / 22  & 0.004 / 0.973 / 21 \\ 
Gradient Boosting (N=100) & 0.006 / 0.955 / 35  & 0.005 / 0.978 / 17  & 0.004 / 0.974 / 20  & 0.004 / 0.981 / 15 \\ 
Gradient Boosting (N=200) & 0.005 / 0.956 / 34  & 0.004 / 0.983 / 13  & 0.004 / 0.977 / 18  & 0.004 / 0.983 / 13 \\ 
AdaBoost (N=25) & 0.008 / 0.922 / 61  & 0.006 / 0.951 / 38  & 0.005 / 0.951 / 38  & 0.005 / 0.967 / 26 \\ 
AdaBoost (N=50) & 0.008 / 0.923 / 60  & 0.006 / 0.962 / 30  & 0.005 / 0.955 / 35  & 0.004 / 0.960 / 31 \\ 
AdaBoost (N=100) & 0.008 / 0.923 / 60  & 0.006 / 0.964 / 28  & 0.006 / 0.959 / 32  & 0.004 / 0.963 / 29 \\ 
AdaBoost (N=200) & 0.008 / 0.923 / 60  & 0.005 / 0.963 / 29  & 0.005 / 0.958 / 33  & 0.004 / 0.964 / 28 \\ 
SVR (RBF kernel) & 0.008 / 0.936 / 50  & 0.006 / 0.967 / 26  & 0.004 / 0.977 / 18  & 0.004 / 0.978 / 17 \\ 
SVR (Polynomial kernel) & 0.010 / 0.937 / 49  & 0.005 / 0.964 / 28  & 0.005 / 0.965 / 27  & 0.005 / 0.968 / 25 \\ 
SVR (Linear kernel) & 0.013 / 0.921 / 62  & 0.008 / 0.950 / 39  & 0.007 / 0.969 / 24  & 0.006 / 0.965 / 27 \\ 
Random Forest (N=25) & 0.006 / 0.946 / 42  & 0.004 / 0.978 / 17  & 0.004 / 0.972 / 22  & 0.004 / 0.976 / 19 \\ 
Random Forest (N=50) & 0.006 / 0.953 / 37  & 0.004 / 0.976 / 19  & 0.004 / 0.969 / 24  & 0.004 / 0.977 / 18 \\ 
Random Forest (N=100) & 0.006 / 0.945 / 43  & 0.004 / 0.978 / 17  & 0.004 / 0.974 / 20  & 0.004 / 0.977 / 18 \\ 
Random Forest (N=200) & 0.006 / 0.951 / 38  & 0.004 / 0.976 / 19  & 0.005 / 0.972 / 22  & 0.004 / 0.978 / 17 \\ 

\hline
\end{tabular}
}
\caption[]{The new baseline, with improvements caused by all advances from the old baseline.}
\label{tbl:newBaseline}
\end{table*}

\begin{table*}[]
\centering
\resizebox{0.83\linewidth}{!}{%
\begin{tabular}{l||c|c|c|c}
\hline
\hline
 &  \multicolumn{4}{c}{\textbf{MAE / Monotonicity Score / \#Monotonicity Violations}} \\
 \hline
 & 100.0\% acceleration & 96.7\% acceleration & 93.3\% acceleration & 90.0\% acceleration\\ 
\textbf{Algorithm} & (0 epochs) & (3 epochs) & (6 epochs) & (9 epochs)\\ 
\hline
Linear Regression & 0.028 / 0.832 / 32  & 0.023 / 0.905 / 18  & 0.018 / 0.926 / 14  & 0.011 / 0.926 / 14 \\ 
Linear Regression (D=0.5) & 0.026 / 0.837 / 31  & 0.018 / 0.837 / 31  & 0.013 / 0.879 / 23  & 0.008 / 0.937 / 12 \\ 
Linear Regression (D=0.25) & 0.021 / 0.842 / 30  & 0.013 / 0.911 / 17  & 0.008 / 0.932 / 13  & 0.006 / 0.942 / 11 \\ 
Linear Regression (D=2) & 0.029 / 0.832 / 32  & 0.024 / 0.905 / 18  & 0.019 / 0.926 / 14  & 0.013 / 0.926 / 14 \\ 
Linear Regression (Exp) & 0.030 / 0.832 / 32  & 0.025 / 0.889 / 21  & 0.020 / 0.926 / 14  & 0.011 / 0.916 / 16 \\ 
Linear Regression (Log) & 0.027 / 0.837 / 31  & 0.021 / 0.905 / 18  & 0.015 / 0.916 / 16  & 0.009 / 0.932 / 13 \\ 
Linear Regression (Sigmoid) & 0.026 / 0.837 / 31  & 0.018 / 0.837 / 31  & 0.013 / 0.874 / 24  & 0.009 / 0.942 / 11 \\ 
SVR (Polynomial kernel) & 0.017 / 0.816 / 35  & 0.027 / 0.758 / 46  & 0.027 / 0.832 / 32  & 0.031 / 0.805 / 37 \\ 
SVR (Linear kernel) & 0.027 / 0.832 / 32  & 0.021 / 0.863 / 26  & 0.018 / 0.911 / 17  & 0.013 / 0.932 / 13 \\ 

\hline
\end{tabular}
}
\caption[]{Left extrapolation performances.}
\label{tbl:leftExtrapolation}
\end{table*}

\begin{table*}[]
\centering
\resizebox{0.83\linewidth}{!}{%
\begin{tabular}{l||c|c|c|c}
\hline
\hline
 &  \multicolumn{4}{c}{\textbf{MAE / Monotonicity Score / \#Monotonicity Violations}} \\
 \hline
 & 100.0\% acceleration & 96.7\% acceleration & 93.3\% acceleration & 90.0\% acceleration\\ 
\textbf{Algorithm} & (0 epochs) & (3 epochs) & (6 epochs) & (9 epochs)\\ 
\hline
Linear Regression & 0.015 / 0.847 / 29  & 0.009 / 0.826 / 33  & 0.008 / 0.858 / 27  & 0.007 / 0.863 / 26 \\ 
Linear Regression (D=0.5) & 0.014 / 0.853 / 28  & 0.009 / 0.842 / 30  & 0.007 / 0.847 / 29  & 0.007 / 0.853 / 28 \\ 
Linear Regression (D=0.25) & 0.011 / 0.853 / 28  & 0.014 / 0.832 / 32  & 0.013 / 0.826 / 33  & 0.014 / 0.816 / 35 \\ 
Linear Regression (D=2) & 0.017 / 0.853 / 28  & 0.010 / 0.826 / 33  & 0.009 / 0.837 / 31  & 0.007 / 0.863 / 26 \\ 
Linear Regression (Exp) & 0.019 / 0.847 / 29  & 0.011 / 0.816 / 35  & 0.009 / 0.811 / 36  & 0.008 / 0.847 / 29 \\ 
Linear Regression (Log) & 0.014 / 0.853 / 28  & 0.008 / 0.842 / 30  & 0.007 / 0.853 / 28  & 0.007 / 0.858 / 27 \\ 
Linear Regression (Sigmoid) & 0.014 / 0.853 / 28  & 0.008 / 0.842 / 30  & 0.007 / 0.853 / 28  & 0.007 / 0.863 / 26 \\ 
SVR (Polynomial kernel) & 0.018 / 0.879 / 23  & 0.017 / 0.879 / 23  & 0.015 / 0.842 / 30  & 0.018 / 0.837 / 31 \\ 
SVR (Linear kernel) & 0.016 / 0.853 / 28  & 0.009 / 0.837 / 31  & 0.008 / 0.832 / 32  & 0.007 / 0.842 / 30 \\ 

\hline
\end{tabular}
}
\caption[]{Right extrapolation performances.}
\label{tbl:rightExtrapolation}
\end{table*}

\begin{table*}[]
\centering
\resizebox{0.83\linewidth}{!}{%
\begin{tabular}{l||c|c|c|c}
\hline
\hline
 &  \multicolumn{4}{c}{\textbf{MAE / Monotonicity Score / \#Monotonicity Violations}} \\
 \hline
 & 100.0\% acceleration & 96.7\% acceleration & 93.3\% acceleration & 90.0\% acceleration\\ 
\textbf{Algorithm} & (0 epochs) & (3 epochs) & (6 epochs) & (9 epochs)\\ 
\hline
Linear Regression & 0.033 / 0.958 / 8  & 0.024 / 0.979 / 4  & 0.015 / 0.947 / 10  & 0.013 / 0.932 / 13 \\ 
Linear Regression (D=0.5) & 0.033 / 0.963 / 7  & 0.023 / 0.979 / 4  & 0.013 / 0.937 / 12  & 0.011 / 0.932 / 13 \\ 
Linear Regression (D=0.25) & 0.032 / 0.963 / 7  & 0.017 / 0.968 / 6  & 0.014 / 0.968 / 6  & 0.009 / 0.942 / 11 \\ 
Linear Regression (D=2) & 0.033 / 0.963 / 7  & 0.028 / 0.989 / 2  & 0.018 / 0.963 / 7  & 0.016 / 0.953 / 9 \\ 
Linear Regression (Exp) & 0.034 / 0.963 / 7  & 0.021 / 0.963 / 7  & 0.016 / 0.947 / 10  & 0.014 / 0.926 / 14 \\ 
Linear Regression (Log) & 0.033 / 0.958 / 8  & 0.020 / 0.963 / 7  & 0.014 / 0.937 / 12  & 0.012 / 0.942 / 11 \\ 
Linear Regression (Sigmoid) & 0.033 / 0.963 / 7  & 0.023 / 0.974 / 5  & 0.014 / 0.947 / 10  & 0.011 / 0.926 / 14 \\ 
SVR (Polynomial kernel) & 0.023 / 0.858 / 27  & 0.013 / 0.905 / 18  & 0.017 / 0.932 / 13  & 0.013 / 0.921 / 15 \\ 
SVR (Linear kernel) & 0.029 / 0.953 / 9  & 0.021 / 0.958 / 8  & 0.016 / 0.947 / 10  & 0.012 / 0.963 / 7 \\ 

\hline
\end{tabular}
}
\caption[]{Dual extrapolation performances.}
\label{tbl:dualExtrapolation}
\end{table*}

\section{Experimental Results}
The new performance baseline is reported in Table~\ref{tbl:newBaseline}. It includes all the advances that were explained in Section~\ref{sec:advancements}. As an ablation study, Table~\ref{tbl:allFeaturesVsBestFeatures} reports the performances of selected algorithms with the feature selection engine turned off and on. It demonstrates a dramatic improvement caused by implementing feature selection, especially when the number of features grows. Figure~\ref{fig:oldVsNewBaseline} demonstrates the improvements visually. Tables~\ref{tbl:leftExtrapolation},\ref{tbl:rightExtrapolation},\ref{tbl:dualExtrapolation} report the performances on the left, right and dual extrapolation problems, respectively. It is important to  mention that the absolute number of monotonicity violations is not on the same scale when comparing the regular baseline to any of the extrapolation baselines. That is because the regular baseline contains 40 test samples, while each of the extrapolation baselines contains 20 test samples, whereas the maximal number of monotonicity violations is the number of pairs, $\binom{N}{2}$. In addition, the test set of the dual extrapolation baseline consists of two separate sets of 10 samples each, while there is a notable gap between the two sets. This gap imposes a regression problem that is harder in terms of accuracy, but is easier in terms of maintaining monotonicity, especially when comparing a pair of samples from the two separate sets. Figure~\ref{fig:extrapolations} visualizes the best performances achieved on the extrapolation splits.

The results reported in Table~\ref{tbl:allFeaturesVsBestFeatures},\ref{tbl:newBaseline} imply that when the number of features grows, implementing feature selection becomes more essential. On the other hand, they also imply that under the same conditions, finding the optimal solution becomes harder, because of the high dimensionality, as discussed in Section~\ref{sec:featureSelection}. 

In addition to the new baselines, we report the features selected by the most accurate regression models. Figure~\ref{fig:selectedFeaturesScheme} relates to regressors that were only trained on the scheme features. For each regression algorithm out of a selected group of algorithms, it reports the rate of trained models that selected each of the features, out of the 8\% top performing models of the given regression algorithm. It demonstrates the high importance of one of the new scheme features we introduced in NAAP-440e, which deals with lost receptive field. Similarly, Figure~\ref{fig:selectedFeaturesSchemeAndQuant3} reports the same information when training the regression models on scheme features and features from the first 3 training epochs. It implies that the architecture training losses are more preferable features for accuracy prediction, in comparison to intermediate test accuracies.

\section{Conclusions and Future Work}
We have presented an extension for the NAAP-440 dataset and baseline with a dramatic improvement, caused by (1) adding new tabular features that were mined from the schemes, and (2) applying feature selection. The extensions improved the old baseline from 25 to 13 monotonicity violations, while maintaining the same mean absolute error and even requiring 3x shorter training processes of the candidate architectures, compared to the old baseline's best performances. We presented the most frequent features chosen by the most fit regressors, to get a sense of what features may be essential for the task of neural architecture accuracy prediction from a tabular dataset. In addition, we presented 3 new dataset splits, which impose 3 new extrapolation problems. By that, we created 3 additional baselines for the relatively difficult problem of extrapolation.

In the future, we may try to address the extrapolation problem by improving the baselines presented in this work. As models that can extrapolate strongly rely on the quality and nature of the input features, we may apply non-linear operators on the tabular features and develop a method for wisely creating and selecting compound features, which will be fed to the regression models.

{\small
\bibliographystyle{ieee_fullname}
\bibliography{egbib}
}

\end{document}